\documentclass[runningheads]{llncs}
\usepackage{graphicx}
\usepackage{rsfso}
\usepackage{enumitem}
\usepackage{soul}
\usepackage{bera}% optional: just to have a nice mono-spaced font
\usepackage{listings}
\usepackage{xcolor}
\usepackage{caption}
\usepackage{subcaption}

\colorlet{punct}{red!60!black}
\definecolor{background}{HTML}{EEEEEE}
\definecolor{delim}{RGB}{20,105,176}
\colorlet{numb}{magenta!60!black}

\lstdefinelanguage{json}{
    basicstyle=\normalfont\ttfamily,
    numbers=left,
    numberstyle=\scriptsize,
    stepnumber=1,
    numbersep=8pt,
    showstringspaces=false,
    breaklines=true,
    frame=lines,
    backgroundcolor=\color{background},
    literate=
     *{0}{{{\color{numb}0}}}{1}
      {1}{{{\color{numb}1}}}{1}
      {2}{{{\color{numb}2}}}{1}
      {3}{{{\color{numb}3}}}{1}
      {4}{{{\color{numb}4}}}{1}
      {5}{{{\color{numb}5}}}{1}
      {6}{{{\color{numb}6}}}{1}
      {7}{{{\color{numb}7}}}{1}
      {8}{{{\color{numb}8}}}{1}
      {9}{{{\color{numb}9}}}{1}
      {:}{{{\color{punct}{:}}}}{1}
      {,}{{{\color{punct}{,}}}}{1}
      {\{}{{{\color{delim}{\{}}}}{1}
      {\}}{{{\color{delim}{\}}}}}{1}
      {[}{{{\color{delim}{[}}}}{1}
      {]}{{{\color{delim}{]}}}}{1},
}

\begin{document}
\title{Decomposed Inductive Procedure Learning}
% an abbreviated paper title herehttps://www.overleaf.com/project/5f74c9c65a4f350001629bf5
%
%First Author\inst{1}\orcidID{0000-1111-2222-3333}
\author{Daniel Weitekamp\inst{1} \and
       Christopher MacLellan\inst{2} \and
       Erik Harpstead\inst{1} \and
       Kenneth Koedinger\inst{1}
        }
\authorrunning{D. Weitekamp et al.}
% If there are more than two authors, 'et al.' is used.https://www.overleaf.com/project/5e7a083b155e280001f3924e

\institute{Carnegie Mellon University, Pittsburgh, PA 15213, USA \and
           Drexel University, Philadelphia, PA 19104, USA}
\maketitle              % typeset the header of the contribution
\begin{abstract}
Recent advances in machine learning have made it possible to train artificially intelligent agents that perform with super-human accuracy on a great diversity of complex tasks. However, the process of training these capabilities often necessitates millions of annotated examples---far more than humans typically need in order to achieve a passing level of mastery on similar tasks. Thus, while contemporary methods in machine learning can produce agents that exhibit super-human performance, their rate of learning per opportunity in many domains is decidedly lower than human-learning. In this work we formalize a theory of Decomposed Inductive Procedure Learning (DIPL) that outlines how different forms of inductive symbolic learning can be used in combination to build agents that learn educationally relevant tasks such as mathematical, and scientific procedures, at a rate similar to human learners. We motivate the construction of this theory along Marr's concepts of the computational, algorithmic, and implementation levels of cognitive modeling, and outline at the computational-level six learning capacities that must be achieved to accurately model human learning. We demonstrate that agents built along the DIPL theory are amenable to satisfying these capacities, and demonstrate, both empirically and theoretically, that DIPL enables the creation of agents that exhibit human-like learning performance.   
\end{abstract}
%
%
% \section \subsection \subsubsection \paragraph

\section{Introduction} \label{intro}
The engineering of machines that possess human-like intelligence is perhaps the greatest challenge among computer scientists of the 21st century. To date, considerable strides have been made toward the creation of artificial intelligence capable of human-level faculties of visual, logical, and linguistic reasoning \cite{silver2016mastering} \cite{voulodimos2018deep} \cite{devlin2018bert}. The successes of the fields of machine learning and artificial intelligence, however, have not, for the most part, been accompanied by a serious analysis of artificial intelligence's capability to replicate the efficiency of human learning, on educationally relevant tasks such as language acquisition, reading, writing, math and STEM problem solving. To this end, modern machine learning approaches often fall short not in their capabilities to achieve human-level performance, but in their capacity to be trained to a point of sufficient capability from the handful of rich instructional interactions sufficient to teach humans to the point of content mastery.

\subsection{Theories of Human Learning}
An artificially intelligent agent with the capability to replicate human-level learning gains from the diversity of stimuli that humans are capable of learning from is not only of practical use for the sake of simplifying the automation of tasks in the real world, but is also a necessary innovation for the purposes of developing theories of human learning. Experimental psychology has been a driving force of studying human learning, but just as the study of machine learning has largely fallen short of producing high fidelity theories of human learning because of a focus on replicating only human performance,  experimental psychology has failed to do the same for its focus on using the discovery of "effects" (experimentally reproducible phenomena) as the sole means of investigating cognitive structures. Newell \cite{newell1973you} has described the process of theory generation in experimental psychology as "playing 20 questions with the universe"---questioning whether the sum of answers to increasingly specific binary experimental questions can possibly yield a full theory of the mind. Likewise, van Rooij and Baggio \cite{rooij2020theory} liken the process of trying to construct theories of cognitive capacities from experimental effects to writing a novel from a collection of random sentences. They point out that there are likely infinite numbers of reproducible "effects", but only a small subset of these may clearly shed light on the inner workings of human cognition. Van Rooij and Baggio suggest that rather than relying solely on a bottom up process of theory construction, a constructive \textit{theoretical cycle} can be employed whereby theories of high verisimilitude (a priori plausibility) are constructed before experimentation to satisfy core plausibility requirements. Others including Newell, have advocated for similar approaches \cite{anderson1990adaptive} \cite{cummins2000does} \cite{newell1973you}.

In considering the verisimilitude of a cognitive theory it is helpful to consider Marr's three levels by which any theory of cognitive capacity may be understood \cite{marr1982vision}, as each level may impose different constraints on the plausibility of a cognitive theory:

\begin{enumerate}
    \item \textbf{computational-level theory}:  Specifies the general nature of the inputs and outputs of the function defining the capacity and the general logic of the strategy by which the capacity is carried out.
    %quote: What is the goal of the computation, why is it appropriate, and what is the logic of the strategy by which it can be carried out?
    \item \textbf{algorithmic-level theory}: Specifies the algorithm by which the capacity is carried out, including a specification of the representation of the inputs and outputs. 
    % quote: How can this computational theory be implemented? In particular, what is the representation for the input and output, and what is the algorithm for that transformation?
    \item \textbf{implementational-level theory}: Specifies the means by which the capacity is physically realized. 
    
    %quote: How can the representation and algorithm be realized physically?
    
\end{enumerate}

Importantly, any cognitive theory that considers at least the first two levels can be built and embodied in simulation, providing a further empirical means for theory testing. Even before consideration of reproducing known experimental effects, the execution of a cognitive theory embodied in simulation imposes its own plausibility requirements at the computational-level beyond what can be imposed only theoretically. These broadly include the requirement that the simulation succeeds in doing what it was designed to do, which is not something that can always be known a priori, and that it do so within some human-like parameters. 
 
%Talk about Marr
%Talk about vermisilitude 
%Talk about explanatory theory vs. searching for effects
%Explanation sits between the tension of having an explainable theory (20 questions with nature) and producing an explainable learned model (explainable vs performant AI models).
%Maybe more on comp theories of learning... (SOAR, ICARUS, ACT-R?, others? [check Chris's thesis for more refs]).

We suggest the following six criteria, or capacities in Marr's terms, as a contribution toward a computational-level theory of human learning. Further specification of these capacities at the the algorithmic-level may produce novel predictions that can be tested empirically against human learning data. \footnote{An implementation-level (i.e. neural level) theory is of third order concern.}
%An implementation-level theory and simulation of these capacities via complete neural simulation is of second-order concern.
%These capacities should be established theoretically, through specification at the algorithmic and implementation level, and empirically, through matching theoretical predictions with human learning data.

\begin{enumerate}[label=C.{\arabic*}]
    \item \label{C:perf} Through instruction learn to \textit{perform} domain tasks as accurately/inaccurately as human learners.
    \item \label{C:lrnrate} Learn to achieve levels of performance similar to human learners at the same \textit{rate} of performance improvement per learning opportunity as humans.
    \item \label{C:mat} Acquire domain knowledge by interacting with \textit{similar materials} to those used by human learners.
    \item \label{C:inst} Learn from the same kinds of \textit{instructional interactions} that humans can learn from.
    \item \label{C:errors} Model the systematicity and variety of human \textit{errors}.%Produce errors of the same nature as human learners at the same relative frequency as human learners.
    \item \label{C:meth} Employ, under plausible constraints, the same general \textit{methods of learning} as human learners including at least a) deductive reasoning (reasoning from known principles) and b) inductive reasoning (reasoning from observations).\footnote{For the sake of argument we bundle abductive reasoning (assuming the most likely explanation) with inductive reasoning, and note that we are not invoking here the formal logic distinction between deduction and induction which requires in both cases a set of premises.}  
    
\end{enumerate}

\subsection{Structure of This Paper}%Decomposed Inductive Procedure Learning (DIPL)}

In this work we define the theory of Decomposed Inductive Procedure Learning (DIPL) which outlines how different forms of planning and symbolic inductive machine learning, each exhibiting different forms of inductive bias, can be used in combination to learn complex procedural skills. We demonstrate theoretically that the learning of procedural tasks in a user interface can be simplified by breaking the process of inducing and refining reusable skills into three or more separate but inter-connected learning mechanisms.

The DIPL theory formalizes 16 years of work on simulated learner systems such as SimStudent \cite{matsuda2015teaching} and the Apprentice Learner (AL) Architecture \cite{maclellan2016apprentice}. Simulated learners are artificially intelligent agents that can learn to perform academic tasks, like solving math problems or generating language translations. These tasks are often the targets of Intelligent Tutoring Systems (ITSs), a highly effective and adaptive class of educational technologies \cite{ma2014intelligent}. As such, these simulated learners experience instruction as a form of tutoring whereby they learn, as tutored humans do, from demonstrated examples and from feedback on their attempts to perform these tasks (\ref{C:inst},\ref{C:meth}.b). These demonstration or feedback-based learning opportunities can be provided to the simulated learner either by the ITS itself or by a human instructor (\ref{C:mat}). Section \ref{procedure} further elaborates on our assumptions with regard to ITSs as a training context for simulated learners.

Section \ref{DIPL} specifies the various learning mechanisms in the DIPL theory at the computational-level, and discusses several algorithmic-level commitments that have been made by different simulated learner implementations in the literature. In section \ref{results} we demonstrate empirically that DIPL based simulated learners are well suited to learning ITS tasks at human-level efficiency, that is, with a dozen or so learning opportunities. We show that Apprentice Learner (AL) agents embodying the DIPL theory master two multi-step mathematics tasks with orders of magnitude fewer learning opportunities than a connectionist reinforcement learning method, a symbolic decision tree method, and an AL agent that diverges from the DIPL computational-level theory by using a single learning mechanism to serve the role of two separate mechanisms in the DIPL theory (\ref{C:perf}, \ref{C:lrnrate}). 

Finally, Section \ref{theory} introduces a number of detailed theoretical claims supporting the particular way that DIPL breaks up the process of procedural task learning. Additionally, we demonstrate throughout that the DIPL approach can effectively learn from the diverse kinds of instructional interactions that students can learn from (\ref{C:inst}). With regard to capacity \ref{C:errors}, we invite the reader to refer to prior work on the types of errors simulated learners make during the learning process \cite{weitekamp2020investigating}. 

%this work several means by which the decomposed and typically explainable learning mechanisms used in the DIPL theory are amenable to being trained from the diverse set of instructional interactions (\ref{C:inst}) typical in human-to-human tutoring. %Finally, we show how the inclusion of these interactions can make inductive learning more tractable and efficient (\ref{C:perf}, \ref{C:lrnrate}) without the need for an explicitly defined domain model.

\section{Related Work} \label{related}
\subsection{Extant Cognitive Architectures and Deductive Methods}

A large number of \textit{cognitive architectures} have sought to model the cognitive processes associated with performing and learning tasks. Notable cognitive architectures such as SOAR \cite{laird2019soar} and ACT-R \cite{anderson1997act} are architecturally built around production rule systems, and share some commonalities in their internal structures that have been recently been termed "A Standard Model of the Mind"\cite{laird2017standard}. 

DIPL is particularly focused on the procedural learning component of the standard model.  It elaborates on the standard model's assumption that "procedural composition yields behavioral automatization"  (p. 24).  In DIPL, procedural composition does not merely speed up existing capabilities, but it can also induce new capabilities from examples and feedback.  
Unlike DIPL systems (AL and SimStudent) which learn primarily from positive and negative examples, standard model systems (SOAR and ACT-R) tend to learn from text-based instructions stored in declarative memory that are interpreted by domain-general procedures which are then composed into domain-specific procedures.  A common research objective of these cognitive architectures has been to test theories of the structure and timing of internal cognitive structures by predicting how human subjects' task execution times shorten as they engage in repeated practice \cite{ritter2007cognitive}. With DIPL, we focus on modeling how learning changes error rate, rather than timing.

Historically, cognitive architectures like SOAR and ACT-R have employed a diverse set of learning mechanisms. One frequently used mechanism is "chunking" \cite{laird1986chunking}  or "production-compilation" \cite{taatgen2003production}, whereby sequences of known production rules and other internal knowledge structures are joined into reusable chunks. One common use of chunking in SOAR is a form of goal directed planning called "speed-up learning" where a sequence of production-rules is found to satisfies some preset goal, and is then combined for more immediate use in subsequent tasks. Although directed toward solving a particular instance of a task, "speed-up learning" is decidedly a deductive learning method (\ref{C:meth}.a) since the reasoning it employs is over prior domain knowledge, and the knowledge acquired by this reasoning is within the deductive closure (the space of all things that can be deduced)\cite{soar1987knowledge} from what is already known. 

SOAR and ACT-R have also historically utilized some inductive learning methods as well. Although these methods typically fall short of inducing, as we do in this work, fully formed production rules without explicitly programmed domain-specific knowledge. For example in SOAR, chunking has been directed toward hierarchical percept recognition \cite{soar1987knowledge}. Additionally there have been many cases of reinforcement learning being employed with SOAR and ACT-R to tweak the utilities of domain-specific production rules in the interest of optimizing the execution of various tasks\cite{nason2005soar}\cite{stocco2018biologically}\cite{shapiro2001using}, and in some cases correcting existing domain knowledge \cite{pearson1995toward}.

Historically, work on simulated learners has diverged from these lines of work in the interest of modeling how humans induce domain specific knowledge directly from various forms of instruction provided by a tutoring system or human tutor. Simulated learner research has sought to build agents capable of learning a wide variety of academically relevant topics. In contrast to research utilizing ACT-R and SOAR, research on simulated learners has been primarily concerned with the cognitive processes underlying the creation and refinement of domain-specific skills in response to various forms of instruction typical in a tutoring setting (\ref{C:inst}). Some key considerations are that simulated learners learn incrementally, make mistakes (\ref{C:errors}), and adjust their acquired skills (i.e. production-rules) in response to feedback (\ref{C:meth}.b) provided by an external source.

\subsection{Machine Learning and Program Induction}

In contrast to research on cognitive architectures like ACR-R and SOAR, the field of machine learning, is primarily concerned with automation, and offers a wide variety of inductive methods whereby a general function, program, or concept is produced to explain a set of examples without need for precoded domain knowledge. The study of inductive machine learning methods has made considerable progress toward achieving greater performance on tasks (\ref{C:perf}) in complex environments, however many modern methods dispense with the objective of learning from few examples (\ref{C:lrnrate}) entirely, requiring examples numbering in the thousands, millions or more. By contrast, despite sources of inefficiency---distraction, forgetting, and misunderstanding to name a few---humans learn remarkably quickly, requiring on the order of 5 to 20 examples and practice problems in order to sufficiently master simple skills \cite{ritter2007cognitive}. Some machine learning algorithms have succeeded in significantly reducing the number of examples needed to inductively perform well at classification \cite{vinyals2016matching} and control \cite{duan2017one} tasks. However, the machine learning community has not made significant progress toward the goal of building agents capable of learning from the limited number of instructional opportunities humans typically need to master mathematical, scientific, and linguistic skills. Tackling this problem is no straight forward task.

No single inductive machine learning algorithm comes close to satisfying even a subset of our criteria, and many have further practical drawbacks\footnote{For example, many neural methods that make claims about zero and one-shot learning require considerable pre-training on similar tasks}. Among existing methods innumerable trade-offs exist in terms of performance (\ref{C:perf}), computational efficiency, explainability, and number of required training examples (\ref{C:lrnrate}) \cite{blackbox} \cite{caruana2006empirical}. Within inductive methods a distinction can be drawn between symbolic and vector-distributed or connectionist methods \cite{quinlan1994comparing}. Connectionist methods represent knowledge as weights on connections between nodes. Following improvements in computing hardware and innovations in the backpropagation learning algorithm \cite{rumelhart1985learning}, connectionist deep-learning methods have produced performance advances across a large number of machine learning tasks. By contrast inductive symbolic methods such as decision trees \cite{quinlan_C45}, FOIL \cite{quinlan_FOIL}, and version spaces \cite{mitchell1982generalization}, are often less accurate than connectionist methods in complex or noisy domains, but can produce explainable knowledge in the form of logical formulas from relatively small datasets.

A handful of systems employ a combination of inductive and deductive machine learning methods (\ref{C:meth}). Ur and VanLehn \cite{ur1995steps} implemented a simulated student that uses the deductive method of explanation based learning (EBL) \cite{dejong1986explanation} whereby theorem proving methods are employed over domain specific knowledge to explain given examples, and then inductively refine theorems in response to positive and negative feedback from a tutor. This method although employing some induction is still grounded in precoded domain knowledge, which raises the question of how domain knowledge can be acquired by a learning agent without need for explicit coding.

\subsection{Inducing Domain-Specific Knowledge}

Wang \cite{wang1995learning} offers one early method for acquiring domain-specific production rules purely inductively, although this methods has some notable limitations including being constrained to the fairly explicitly defined predicate list worlds typical of symbolic planning environments and being limited to conjunctive preconditions. Simulated learner systems overcome these limitations. 

Simstudent \cite{matsuda2015teaching} and the Apprentice Learner (AL) \cite{maclellan2016apprentice} framework are two simulated learner systems that inductively acquire domain skills by working in intelligent tutoring systems (ITSs) built for human students. These two systems, which are the subject of this work, construct procedural domain knowledge (i.e. production rules or skills) inductively using a collection of several interconnected learning mechanisms. By contrast to systems that model just the statistics of errors made by learners \cite{cen2006learning}, simulated learners are best described as computational theories \cite{maclellan2017computational} or computational models \cite{matsuda2014investigating}\cite{weitekamp2019toward} of human learning. They embody a theory of precisely how knowledge is induced and refined in response to instructional interactions and how that acquired knowledge produces particular correct and incorrect actions in response to particular stimuli. 

Simulated learners have also been used to facilitate a form of programming by demonstration where a human user can tutor the simulated learner in an academically relevant task for the purposes of authoring the grading behavior of an intelligent tutoring system \cite{matsuda2015teaching} \cite{weitekamp2020CHI}. This authored behavior can be exported as a set of human-readable rules that can be executed independently of the agent. In this work, however, we will primarily discuss the use of simulated learner models that learn directly from existing intelligent tutoring systems.
\section{Procedural Tasks in Intelligent Tutoring Systems} \label{procedure}

Intelligent Tutoring Systems (ITSs) automate elements of human tutoring and, more generally, strive to support an optimal student learning experience \cite{vanlehn2006behavior}.
%bridge the gap between human-to-human instructional interactions and human-to-computer interactions by providing a canonical format into which known tutoring interactions can be automated
Cognitive Tutors \cite{ritter2007cognitive} are a particular form of ITS that applies human learning research and is particularly focused on supporting the acquisition of procedural skills. As such, these tutors track student problem solving as it is revealed in multi-step user interfaces where, for example, students enter lines of an equation solution or digits in a long form arithmetic solution \cite{anderson1995cognitive}. Cognitive Tutors provides students with step-by-step correctness feedback and hints, culminating, if needed, in bottom-out hints that demonstrate the next correct step for the student. These core interactions, used in combination with student knowledge modelling that mitigates over- and under-practice of topics, have produced considerable improvements in student learning compared to baseline instructional strategies in a large diversity of domains \cite{cognitivetutors}\cite{pane2014effectiveness}. For the purposes of constructing theories of human learning, Cognitive Tutors provide learning environments that are both well-defined and ecologically valid (\ref{C:mat}) since a diverse set of topics have been taught to humans with Cognitive Tutors relatively effectively \cite{aleven2016example}.

\subsection{A Characterization of Procedural Task Complexity}\label{complexity}

Procedural tasks are tasks performed in a series of actions. The simplest procedural tasks are completed with a fixed sequence of actions. Several systems have been designed to learn these sorts of tasks from demonstrations, including SUGILITE \cite{li2017sugilite} and SmartEDIT \cite{lau2001learning} to name a few. However, procedural tasks typical in academic domains can exhibit a number of complexity factors that are often not well addressed in prior work on automated procedure learning.  For example in the two domains evaluated in this paper (fraction arithmetic and multi-column addition) the following complexity factors are present: 
%We provide the following non-exhaustive list of complexity factors that we have found to occur in procedural tasks taught by ITSs:

\begin{enumerate}[label=P.{\arabic*}]
\item  \label{pc:var} \textit{Variablized Transformations}: Steps exist that not only require classification (e.g., to determine which menu item to click), but the transformation of novel information in the interface, such as, producing the sum of given numbers or concatenating "ed" to the end of a given verb. 
%These steps require copying and transforming unknown content  For example, a step that copies a value from one box to the next is variablized, whereas clicking a button is not. 
\cite{koedinger2012knowledge}
\item  \label{pc:mental} \textit{Presence of Mental Subtasks}: Not every step in the interface can be achieved in a single operation,
requiring some mental information to be computed between steps. 
\item \label{pc:mult} \textit{Multiple Procedures One Interface}: The human or agent applying the procedure has been exposed to multiple procedures that can be done in the same interface, but must learn cues from specific content in the interface to determine which one should be applied.
\item \label{pc:goal} \textit{No Verifiable Goal/Subgoal States}: The goal state is not hard-coded or known in advance of applying the procedure.
\item \label{pc:cont} \textit{Sequence Depends on Context}: Within a single type of procedure the number and type of steps taken depends on the content of the presented task. For example, depending on starting conditions, recursive tasks may require any number of iterations, and the set of correct solution paths to a problem may vary depending on its start state. \cite{koedinger2012knowledge}

\end{enumerate}

It is not uncommon for an academic skill, addressed by a Cognitive Tutor, to exhibit many of these complexity factors at a time. The majority of math and science domains require \ref{pc:var} and \ref{pc:mental}. For instance, the typical procedure for adding large numbers requires repeatedly finding intermediate sums---this exhibits \ref{pc:var} since there are variable arguments to the action. These sums then need to be decomposed into their ones and tens digits, while accounting for the value of sum itself only mentally \ref{pc:mental} (see Figure \ref{fig:mc} in the next section). Additionally, \ref{pc:goal} is present for just about any task that cannot be reduced to a search problem. \ref{pc:cont} includes problems where all problem instances cannot be embedded in the same finite state machine graph (and by extension embedded in an example-tracing based tutoring system \cite{aleven2009new}). Adding large numbers is also an example of this case since the actual steps that need to be taken vary depending on whether the tens digits of intermediate sums need to be carried across columns.

\subsection{Simulated Learners in Two Math Cognitive Tutors}

In this subsection we present two examples of math cognitive tutors. We will refer to these tutors again in section \ref{results}. 

\subsubsection{Fraction Arithmetic}
\begin{figure}
\begin{center}
  \includegraphics[width=0.6\linewidth]{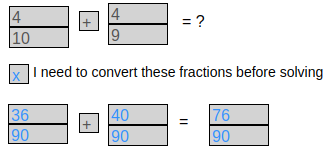}
  \caption{The fraction addition tutoring system completed for the problem $\frac{4}{10} + \frac{4}{9}$. The blue text are steps entered by real or simulated students.}
  \label{fig:fraction_addtion}
\end{center}
\end{figure}

The Fraction Arithmetic tutoring system shown in Figure \ref{fig:fraction_addtion} teaches three different procedures and thus exemplifies the \textit{multiple procedures} (\ref{pc:mult}) complexity factor.  These procedures are: 1) the addition of fractions when the denominator is different, 2) the addition of fractions when the denominator is the same, and 3) multiplication of two fractions. Because only the first of these three requires extra interface elements to convert the fractions to a common denominator, there is a step in the interface to indicate whether or not common denominators need first be found. This task domain exhibits the other task complexity factors except for the \textit{sequence depends on content} (\ref{pc:cont}) factor since the three procedures taught in the interface all have a fixed number and type of steps. Because knowing when a problem is done is an important skill in any procedure, a student or simulated learner must press the "done" button (not shown) to indicate problem completion.

\subsubsection{Multi-Column Addition}
The Multi-Column Addition tutoring system shown in Figure \ref{fig:mc} teaches what is often called the standard or traditional algorithm for adding large numbers. The example shows the three-by-three digit addition problem 539 + 421. Students are required to add the numbers in each column and carry the tens digit of the sum (if it is greater than 10) to the next column if necessary. This domain includes all complexity factors except for \ref{pc:mult}. Unlike the fraction arithmetic domain, \ref{pc:cont} is present in multi-column arithmetic since 1) the instances where a 1 needs to be carried depends on the start state, and 2) the underlying procedure is recursive and works for numbers with any number of digits. 

\begin{figure}
\centering
\includegraphics[width=.5\linewidth]{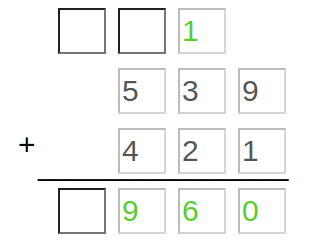}

%\caption{539+421 completed in the tutoring interface}
  \caption{The multi-column addition tutoring system completed for the problem 539 + 421. The green text are steps entered by real or simulated students.}
\end{figure}\label{fig:mc}

\subsection{The interface between ITSs and Simulated Learners}

At every step of an ITS problem, simulated learners 'see' the state of the ITS encoded as a set of interface element features that indicate information such as the value displayed by the interface element, whether the element has been locked---either because it has already been filled or because the interface element is non-interactive---and information about where the interface element falls on the page relative to others interface elements. For example, an interface element may be inside a particular row or column or above, below, or to the left or or right of other interface elements. Below is a slice of a typical state encoded as JSON that an Apprentice Leaner agent would see for a multi-column addition problem:

\begin{lstlisting}[language=json,firstnumber=67]
           ...
           'x': 572.46875,
           'y': 360},
 'inpA1': {'above': 'hidden1',
           'below': 'inpB1',
           'contentEditable': False,
           'dom_class': 'CTATTextInput',
           'height': 40,
           'id': 'inpA1',
           'offsetParent': 'background-initial',
           'to_left': 'inpA2',
           'to_right': '',
           'type': 'TextField',
           'value': '9',
           'width': 40,
           'x': 552.46875,
           'y': 180},
 'inpA2': {'above': 'carry1',
           'below': 'inpB2',
           ...
\end{lstlisting}

Inherent in the use of this symbolic JSON format is the assumption that the agent has as background knowledge the ability to recognize alphanumeric symbols and the presence of positive and negative feedback (via changes of HTML styles), and non-domain-specific spatial relationships like embedding structure and adjacency. In principle these capacities would be fulfilled by additional visual processing and structuring modules. However for the sake of modeling learning in most academic domains, these capacities are already within the capabilities the target learners, and thus out of the scope of the learning that needs to modeled or automated for practical use. This assumption however only extends to domain-general visual processing, and thus simulated learners must still learn any domain specific structure within such a partially parsed representation including for example, in the case of algebra problems, the hierarchical nature underlying the raw text of an algebraic expressions \cite{li2015integrating}.

Simulated Learners within the AL framework expose two general functions train(), and act(). Act() takes in a state and produces an interface action in the form of a Selection-ActionType-Input (SAI) triple. The pieces of the SAI include 1) the selection---the interface element(s) that the agent is acting on, 2) the action type---what the agent is doing to that interface element, and 3) the input---the parameters to the action being performed on the interface element. For example, inserting the text "5" on an interface element called "C" would have an SAI triple of ("C", "UpdateTextField", {"value" : "7"}). Train() is called to trigger learning in response to instructional opportunities including demonstrations of correct actions and reward feedback on a simulated learners' attempted actions. Within an ITS interface demonstrations are typically experienced as bottom-out hints (i.e. a hint that includes the answer to the next step), and reward feedback is typically experienced as a red or green style change. Train() takes at least three arguments, a state, an SAI, and a continuous reward value. Reward values are less than zero for negative feedback and more than zero for positive feedback.

\begin{figure}
    \centering
    \includegraphics[width=.5\linewidth]{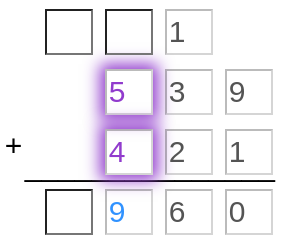}
    \caption{An example of \textit{foci-of-attention}. The tutor or tutoring system has demonstrated placing the value of the intermediate sum 9, and has marked 5 and 4 as \textit{foci-of-attention} indicating that they were used to produce the 9.}
    \label{fig:foci}
\end{figure}

Train() can optionally take additional instructions and feedback from a human instructor, or tutoring system. These additional forms of instruction can take just about any form, so long as the agent is configured to utilize it. One example, is \textit{foci-of-attention}, a set of interface elements whose values are utilized to take a next step in a task. For example, when demonstrating the final addition step of the problem 539+421 (see Figure \ref{fig:foci}), the interface elements with values 5 and 4 would be selected as foci-of-attention to indicate that they were used to obtain 9. We will discuss these additional forms of instruction further in section \ref{DIPL} .

%When an agent produces an SAI in response to a call to act(), this is merely a proposal to the tutoring system to change the state, and the actual state change that occurs in response to an SAI is determined by the tutoring system itself---the inner-working of which are not known to the agent. For this reason skills are not quite like the idea of operators in symbolic planning literature since the actual state change produced by applying a skill is not known to the agent with any certainty. For example, in Cognitive Tutors steps often involve placing values in various text boxes. When an action is correct the value placed in the box turns green and the field is locked (so the student cannot undo a correct step), but if the action is incorrect then the value turns red and remains editable. This visual feedback is important for learning since green and red indicate positive and negative feedback respectively, and for the agent to know where it is in a problem since the locking of the field indicates the completion of a step. 

%State representation, examples, feedback. I.e. what are the affordances of an intelligent tutoring system? They are a subset of the instructional possibilities afforded by a human instructor, but they are surely sufficient at least to completely train a human to master the content it is teaching. Thus, an AI ought to be able to replicate this learning. Since the ITS surely possesses all the information necessary to produce a sufficient solution.%

\section{Decomposed Inductive Procedure Learning (DIPL)} \label{DIPL}

DIPL specifies at the computational-level a theory of how skills are acquired and refined in simulated learners by employing a number of distinct learning mechanisms. DIPL outlines, at the computational-level, the common structure of simulated learners such as SimStudent and various Apprentice Learner (AL) agents that differ in their theoretical commitments at the algorithmic-level. The AL framework, provides modules for each of these various learning mechanisms, and is the primary basis from which the DIPL theory is derived, however, the AL framework can be used to construct agents that do not adhere to the DIPL theory (for example \cite{maclellan2020optimizing}), so DIPL and AL are not completely synonymous.

In defining DIPL here we describe, in far greater detail than has been previously attempted, (a) the boundaries between the computational and algorithmic theoretical commitments made by existing simulated learners like SimStudent and agents built with the AL framework, (b) the variability in algorithmic commitments made by various generations of simulated learner research, and (c) how simulated learners embodying the DIPL theory frame the problem of inductive procedure learning such that it is tractable (\ref{C:perf},\ref{C:lrnrate}), and amenable to diverse forms of instructional intervention (\ref{C:inst}). 

\subsection{Skill Characterization of Procedural Task Learning}

Simulated learners perform procedural tasks by inducing skills that accurately perform steps in a task domain. Skills are implemented as production rules composed of a set of preconditions called the left-hand side (LHS) and a set of effects called the right-hand side (RHS). 
%Conceptually the LHS and RHS of these production rules are equivalent to 'condition' and 'repsonse' in the KLI framework \cite{}.%
When an agent observes a particular state of a tutoring system each of the LHSs of its skills may each match several sets of interface elements on which each skill could be applied. These sets of interface elements are called a \textit{binding}, and include 1) the elements to be used as the arguments to the RHS of the skill (this is analogous to \textit{foci-of-attention} shown in purple in Figure \ref{fig:foci}) and 2) the \textit{selection} (the interface element to be acted upon, blue in Figure \ref{fig:foci}). A skill combined with a particular \textit{binding} is called a \textit{skill application}. When a \textit{skill application} is evaluated the associated skill's RHS is applied on the \textit{binding} to produce an action in the form of an SAI. SAIs are applied to a tutoring system interface to change its state. For a particular state, since a simulated learners' skills may each apply in several ways, the agent often has a \textit{conflict set} of multiple possible next actions. The simulated learner must choose among these conflicting skill applications to decide which action to apply at each state. 

Characterizing procedure learning as a matter of acquiring a number of independent skills is well grounded in existing theories of human learning. Skills, as we refer to them here, have a one-to-one relationship with the idea of Knowledge Components (KCs) \cite{koedinger2012knowledge}, which are independent pieces of procedural or declarative knowledge that must be learned to master a particular domain. Prior work has shown that using SimStudent's skills as KC labels for task steps produced KC attribution models that better predict student data than models designed by experts \cite{li2011machine}. 

\begin{figure}
\begin{center}
  \includegraphics[width=\linewidth]{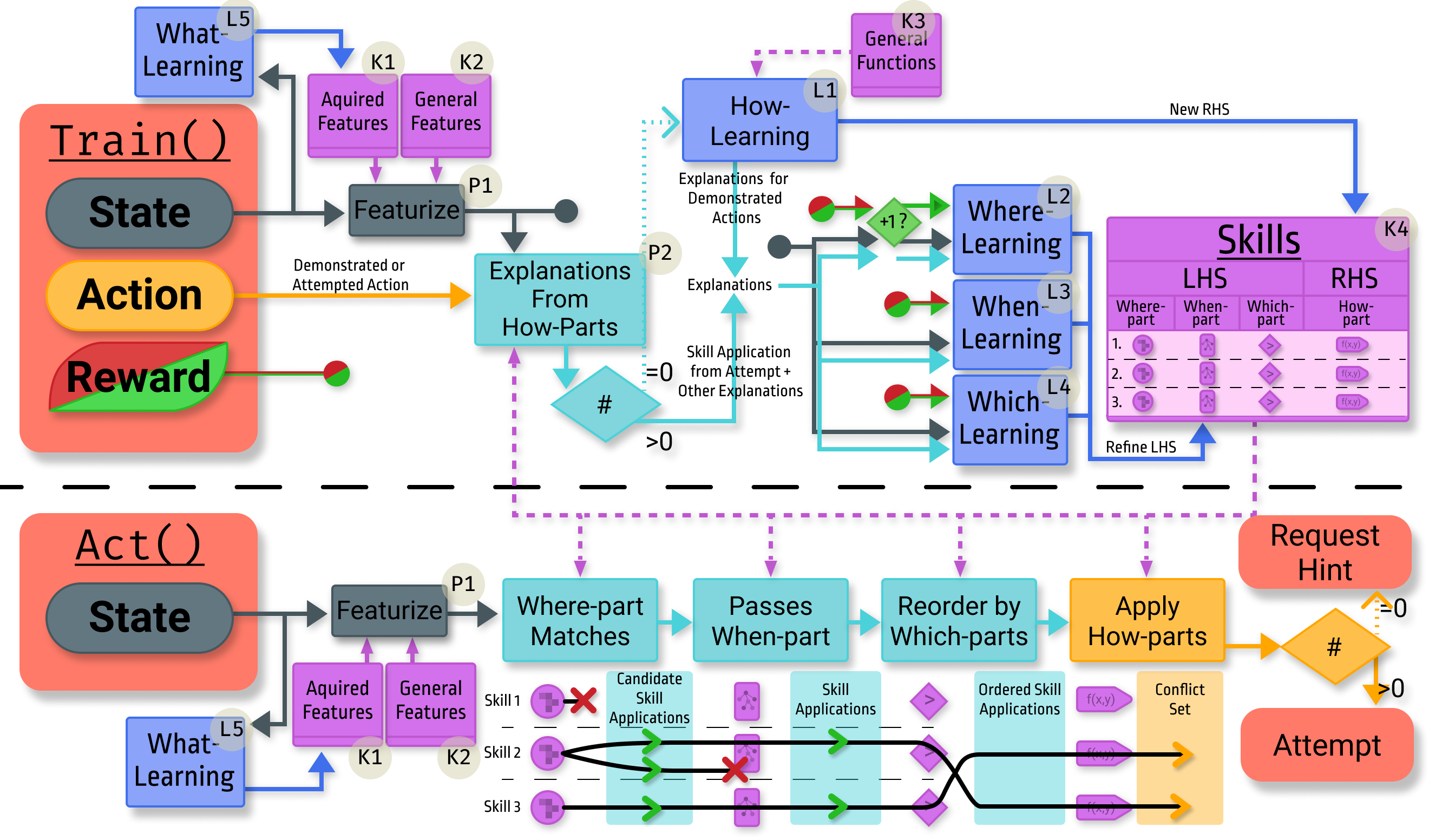}
  \caption{%The Apprentice Learner framework represented in terms of 
  The DIPL computational-level theory. Functions Train() (above the dashes) and Act() (below the dashes) are shown in red. Solid arrows indicate the flow of information such as state (Dark Grey), actions (Gold), explanation and skill applications (Cyan),  reward (Red/Green), and learning results (Blue). Dashed arrows indicate the utilization of various forms of long-term knowledge K1-K4. In both the Train() and Act() functions the observed interface state is augmented with acquired (K1) and general (K2) features (P1). When train receives a demonstrated action the agent attempts to explain it from the \textit{how-parts} of known skills (P2). If no such explanations can be found then \textit{how-learning} (L1) is engaged to build a new explanation from the general functions (K3). A new RHS from this explanation is induced creating a new skill in the skill set (K4). When Train() receives feedback on an attempted action the skill application that produced the action is typically used as the explanation. Once an explanation is found \textit{where-}, \textit{when-}, and \textit{which-learning} (L2-L4) are triggered  to refine the LHS of the explaining skill. They take as input the augmented state, explanations, and reward. \textit{Where-learning} (L2), however, is typically only trained with positive feedback. Finally, the Act() function produces a conflict set of next actions given the augmented state. Act() first applies each skill's \textit{where-part} to find the \textit{candidate skill applications} and then uses the associated \textit{when-parts} to obtain the final set of \textit{skill applications}. Then the \textit{which-parts}  reorder the skill applications and they are evaluated to produce the conflict set of actions. If the conflict is empty then a hint is requested, otherwise the top action on the conflict set is attempted.}
  \label{fig:DIPL_diagram}
\end{center}
\end{figure}

\subsection{Overview of DIPL}

%Apprentice Learner framework at the computational-level in terms the various learning mechanisms described in the DIPL theory. 
%including the structure and function of skills and other types of long-term knowledge (Purple), the role of explanations in the learning process (Cyan), and the five learning mechanisms (Blue).
% experienced within some interface

Figure \ref{fig:DIPL_diagram} outlines the theoretical commitments of the DIPL computational level theory and the various learning mechanisms within it. The function Train() in the upper portion of Figure \ref{fig:DIPL_diagram} represents the learning processes engaged when a simulated learner observes a new instructional opportunity. These processes can be roughly broken into three parts 1) the augmentation of the observed state with new features derived from various forms of knowledge (black), 2) the explanation of demonstrated actions (cyan), and 3) the engagement of various learning mechanisms (blue) to induce or refine skills (purple, K4). 

At the state augmentation step various forms of knowledge add new features to the state. These include \textit{domain-general feature knowledge} and \textit{acquired feature knowledge}. \textit{Domain-general feature knowledge} (K2) includes capacities for adding new non-domain specific features to a state that are assumed to be prior knowledge in a target learning population. For example a \textit{domain-general feature} called Equals(x,y) may be given to an agent to implement a capacity to recognize that two values in an interface are equivalent. \textit{Acquired feature knowledge} (K1) is learned by \textit{what-learning} and typically takes the form of representational features pertaining to the underlying hierarchical structure of a domain---for example the hierarchy of equations, expressions, terms, and operations in algebra or the structures of functional groups, atoms, and bonds in molecular diagrams. Prior work with Simstudent implemented \textit{what-learning} \cite{li2015integrating} as an unsupervised learning process. Although we include \textit{what-learning} here for completeness the main claims of this work pertain to the three core mechanisms \textit{how}, \textit{where} and \textit{when}.

The \textit{when-}, \textit{where-} and \textit{which-learning} mechanisms are responsible for learning the preconditions for skills, and they are made aware of which skills' \textit{when-}, \textit{where-} and \textit{which-parts} must be refined via an \textit{explanation} of each observed action. An \textit{explanation} specifies the skill and binding of interface elements used to produce an action. If the agent itself produced an action then typically the skill application used to produce the action is used as the \textit{explanation}. However, if the action was demonstrated to the agent then the agent must produce an \textit{explanation} for the action. The cyan portion of the upper part of Figure \ref{fig:DIPL_diagram} shows how a DIPL style simulated learner typically tries to explain demonstrated actions from the \textit{how-parts} of known skills. If existing knowledge fails to produce an \textit{explanation} for an action then \textit{how-learning} is engaged to produced a new skill to explain the action. \textit{How-learning} searches over combinations of interface element values and compositions of \textit{domain-general functions} (K3) to induce a new \textit{how-part} and \textit{binding} that reproduces the demonstrated action. %\textit{How-learning} is most frequently engaged in the first opportunities of a new domain. 

The function Act() in the lower portion of Figure \ref{fig:DIPL_diagram} represents how skills are utilized to take actions in response to an observed interface state. The \textit{where-part} of each skill is first applied to the state to generate a set of \textit{bindings} to produce a set of \textit{candidate skill applications}, which are then verified by the \textit{when-part} to produce a final set of \textit{skill applications}. Each \textit{skill application} is evaluated to produce one action in the \textit{conflict set}---the set of an agents proposed next actions.

The \textit{which-part} of a skill can ascribe some utility value to a skill applications so that the highest utility skill applications are favored---putting their actions higher in the conflict set.  \textit{Which-learning} is however relatively inessential overall for effective learning. Historically SimStudent has not implemented \textit{which-learning}, and AL has implemented it as simply the proportion of positive reward associated with a skill. Some prior work has shown that \textit{which-learning} can however be useful for de-prioritizing or even eliminating buggy skills that have been induced but have \textit{how-parts} that are not strictly correct \cite{weitekamp2021toward}.

The following subsections will discuss the computational commitments made by the three core learning mechanisms (\textit{how}, \textit{where}, and \textit{when}) within the DIPL theory and the algorithm-level commitments made by various simulated learner implementations within the literature.

\subsection{How-Learning Mechanism}
\subsubsection{Computational-Level}
\textit{How-learning} induces new skills and is typically the first learning mechanism employed by a fresh simulated learner. Simulated learners typically ask for demonstrations of correct behavior when their current skills produce an empty conflict set of next actions. When a demonstration is received and cannot be explained by the RHSs of the agent's current skills,  \textit{how-learning} will be triggered to form a new RHS that explains the demonstration. \textit{How-learning} searches for an explanation for each demonstrated SAI by searching for a composition of domain-general functions that can take values found in the current problem state as arguments and produce the demonstrated SAI. When a function composition is found, a new skill is induced with it as the \textit{how-part} (i.e. the RHS). When no composition can be found \textit{how-learning} \textit{bottom's out}, and a new skill is induced with a \textit{how-part} that applies the observed SAI as a constant. 

For example, if a simulated learner was working in a tutoring interface for simplifying algebraic expressions, and received a demonstration that $2x+2x$ can be simplified to $4x$, then one possible explanation a \textit{how-learning} mechanism could produce for this particular demonstration is that in general we should multiply the two coefficients and append the variable to the result. For this particular problem this explanation, which of course is incorrect in general, would suffice. In subsequent problems, for example $5x+3x$, this explanation would produce an incorrect answer $15x$, which would be marked as incorrect. After receiving another demonstration the \textit{how-learning} mechanism will produce a new explanation to generate a new \textit{how-part} that can explain all available demonstrations so far. 

\subsubsection{Algorithmic-Level}
\textit{How-learning} mechanisms employ search in order to form explanations from a preset corpus of domain-general functions. Typically, a forward chaining search process is used to find a composition of domain-general functions that produce the target SAI from values found in the problem state. This forward chaining search typically only proceeds up to a fixed depth. The particular forward chaining method employed in \textit{how-learning} have differed across different simulated learner implementations. For example Matsuda et. al. \cite{matsuda2015teaching} used iterative deepening search to chain together domain-general functions, Maclellan et. al. \cite{maclellan2016apprentice} used an first-order logic based planner, and Weitekamp et. al. \cite{weitekamp2020CHI} repeatedly applied broadcasting tensor operations. 

%It should be noted that for the functions typically used in \textit{how-learning} the kinds of pure backward chaining approaches common in symbolic planning and theorem proving are intractable. For example to backward chain from the goal value "7" using the "Add(?, ?)" function would produce the infinite set of pairs of numbers that sum to 7. Although it has not been implemented in SimStudent or AL, in principle it is possible to use backward chaining only for functions with finite preimages, in conjunction with a normal forward chaining approach. For example if a "Double(?)" (i.e. $f(x) = 2x$) function was in the set of domain general functions then this could be backward chained because for a goal $y$ there always exists one $x$ such that $f(x) = y$.

\subsubsection{Instructional Affordances}
Since the search space of \textit{how-learning} problems typically grows exponentially with the length of their explanation, the tractability and cognitive plausibility of this learning mechanism can be of some concern. Interactive training methods for simulated learners have often provided a means to address tractability concerns by prompting users to indicate the interface elements that were used as arguments in computing a demonstration, Matsuda et. al. termed these  foci-of-attention \cite{matsuda2015teaching}. For example, for a demonstration of simplifying the expression $4y+5x+3x$ the user might select $5$,$3$, and either of the two $x$s. By explicitly specifying these values, the user significantly reduces the size of the search space, by ruling out explanations that might employ extraneous interface elements. 

Even with \textit{foci-of-attention} there are conceivably situations where the search problems faced by \textit{how-learning} are incredibly large---too large perhaps to be feasibly solved by a human. However, in the context of interactively training simulated learners to be used as ITS authoring tools, it can be helpful for the agent to be able to solve these large search problems since doing so may save the user from the inconvenience of having to hard-code a skill's RHS. In these cases careful optimizations of the search process may be beneficial including only applying commuting functions to every combination of values, but not every permutation, and reducing the set of intermediate values used at each depth by only computing the values for the next depth with the unique values from the previous depth. Since the how-search problem involves a great deal of repeated function application in loops, the greatest performance benefits come from simply using contiguous data structures in a compiled programming language. The most recent implementation of how-learning in AL utilizes numba \cite{numba}, a just-in-time compiler for Python to this end.

Beyond foci-of-attention, \textit{how-learning} can benefit from other forms of instruction. For example, SimStudent requires a skill label at each user demonstration. Skill labelling clarifies the skill that the user or tutoring system intended to demonstrate, allowing the simulated learner to employ \textit{how-learning} that searches for a composition of functions that are consistent over multiple demonstrations. 

Another consideration within \textit{how-learning} is that humans are able to interpret written formulas and written or spoken natural language descriptions of how steps of procedures are performed. In the context of an ITS these may be presented in the form of hints, or via some initial text/lecture instruction. Future work may explore how natural language processing of these descriptions could guide \textit{how-learning}, considerably constraining the set of functions that should be considered and the order that they should be employed. In this case \textit{how-learning} can be interpreted not as just a blind guess and check process, but as a process of using grounded examples to disambiguate the meaning of instructions that may have a variety of plausible interpretations.

\subsection{Where-Learning Mechanism}

\subsubsection{Comutational-Level}
The \textit{where-} and  \textit{when-learning} mechanisms collectively produce the LHS of skills. \textit{Where-learning} produces the \textit{where-part} of the LHS of a skill. The \textit{where-part} is a generative function (usually in the form of a set of conditions) that takes a problem state and picks out potential \textit{bindings} of a skill. Recall a \textit{binding} includes a selection interface element (the interface element that will be acted on) and set of argument interface elements, that can be fed into the \textit{how-part} of the skill to produce an action. The \textit{where-part} can be thought of as a sort of attention mechanism that produces a set of candidate applications of skills, that may or may not be correct to apply on the current problem state. The \textit{where-part} of the LHS is only a collection of heuristic conditions that pick out possibilities, whereas \textit{when-learning} is responsible for learning the \textit{when-part} which evaluates the correctness of these possibilities.

When a skill is first induced a \textit{how-part} and \textit{where-part} are learned that are sufficient to reproduce the demonstrated action. As more demonstrations are encountered \textit{where-learning} generalizes the \textit{where-part} so that it is unconstrained enough to accept the new demonstrations. \textit{Where-learning} is driven primarily by demonstrations. In some circumstances it can utilize positive correctness feedback but typically does not incorporate negative correctness feedback. 

%Of all the preconditions in the LHS the \textit{where-part} has the responsibility of acting as a set of heuristic conditions that pick out potential applications of a skill. Note that a skill that has a  \textit{how-part} (i.e. RHS) and \textit{where-part} are sufficient 

%These \textit{where-part} matching rules pick out the selection, the interface element that will be acted on by the skill, and the arguments, the interface elements on which a skill's RHS will be applied when the skill fires. The two parts of a potential application of a skill are 1) a particular binding a set of \textit{where} matching rules to a set of selection and arguments and 2) the value of the RHS formula evaluated over the arguments. Typically the \textit{where} learning mechanism learns exclusively from  demonstrations and like \textit{how-learning} and utilizes foci-of-attention to identify the arguments intended by the user. 

\subsubsection{Algorithmic-Level}

In AL and SimStudent, \textit{where-learning} is implemented as a process of finding the least general generalization \cite{lgg} across a number of examples. \textit{Where-learning} typically instantiates a set of highly specific \textit{where-part} conditions that will match to the \textit{binding} found in the explanation of an initial positive example. As each new example is encountered and an explanation is found for them, the \textit{where-part} conditions are generalized to match to new \textit{bindings}. Thus \textit{where-learning} is a conservative specific-to-general learning process. As more examples are encountered the \textit{where-part} of a skill tends to propose more and more candidate \textit{bindings}, but it typically avoids overgeneralizing beyond what the examples so far entail.  \footnote{Historically AL and SimStudent's \textit{where-learning} mechanisms have been described as implementing version spaces \cite{mitchell1982generalization} \cite{matsuda2015teaching} \cite{maclellan2016apprentice} \cite{weitekamp2020CHI}. However, least general generalization is a better characterization since version spaces implement both general-to-specific and specific-to-general learning, but \textit{where-learning} has historically only implemented the latter (section \ref{theory} describes why this is the case).}

\begin{figure}
    \centering
    \includegraphics[width=\linewidth]{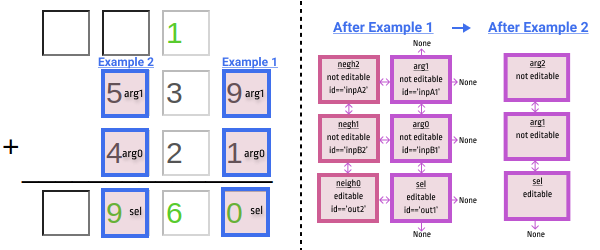}
    \caption{An example of AL's \textit{where-learning} process in the multi-column arithmetic domain. Left: Two demonstrations of the Add2 skill which adds two numbers and places the ones-digit below. Right: After the first demonstration a very constrained \textit{where-part} (purple) is induced which also keeps concepts for neighboring elements (red). After the second demonstration the \textit{where-part} generalizes considerably and several conditions including conditions on the 'id' of the interface elements and conditions on the neighboring elements are dropped. Constraints are preserved keeping the elements vertically aligned and the selection editable.}
    \label{fig:where_learning}
\end{figure}

The the AL \textit{where-learning} mechanism described in \cite{weitekamp2020CHI} generalizes \textit{where-part} matching conditions by performing simple anti-unification. In this \textit{where-learning} mechanism, called 'AntiUnify', a set of  conjunctive conditions are maintained that check for particular features or spatial relationships (i.e. object A must be above, below, to the right, or to the left of object B) between target objects. These conditions are generalized by simply dropping literals in the \textit{where-part} conditions that are not satisfied by new examples of \textit{bindings}. For example, in Figure \ref{fig:where_learning} after one demonstration is observed the \textit{where-part} of the skill LHS is very constrained and matches to just the 3 particular interface elements used in the demonstration. The conditions require the particular 'id' fields of the interface elements, among other constraints. After the second demonstration the \textit{where-part} conditions generalize to the point that they bind to any three vertically aligned interface elements as long as only the bottom one (the selection) is editable\footnote{Recall correct actions lock fields. This change can be conditioned on.}. 

AL also initially keeps around sets of conditions that track neighboring interface elements. In the multi-column addition example shown in Figure \ref{fig:where_learning} these neighboring elements are eliminated upon generalization, however, in some circumstances these neighbor elements can remain and are essential to keeping \textit{where-learning} from over-generalizing. For example, continuing with the example of multi-column addition, for skills that carry '1's across columns, neighboring elements help form an adjacency relationship between the selection and arguments that ensures that the selection can only be in the column to the left of the arguments, instead of in any column. 

Prior implementations of AL agents have used another implementation of \textit{where-learning} called 'MostSpecific' \cite{maclellan2016apprentice} that does not generalize at all, it simply keeps a list of the sets of interface elements demonstrated to it. This simpler \textit{where-learning} mechanism is often sufficient in domains where interface elements remain static, and has the advantage of never over-generalizing. However, this simpler \textit{where-learning} mechanism limits AL's ability to reason about skills across multiple interfaces, or in interfaces that are dynamically generated, like interfaces that are hand drawn or not part of a static page.

Simstudent's \textit{where-learning} mechanism is connected to the hierarchical structure of its working memory, which mirrors the nested list structure of interface elements in a problem display. In Simstudent, \textit{where-learning} is framed as a matter of locating interface elements by a hierarchical retrieval path from the document root to particular interface elements. Simstudent's generalization language is engineered around the JESS rule engine's \textit{multifield} feature \cite{li2012efficient}, so generalizations typically involve altering the matching patterns of lists. For example if SimStudent encounters two demonstrations "row 2, column 3, of table 1", and "row 2, column 4, of table 1" it might generalize to "row 2, any column, of table 1", and perhaps even further to "any row and any column, of table 1" \cite{matsuda2015teaching}. 

\subsubsection{Instructional Affordances}
The bindings found in explanations of demonstrated correct actions are the driving force of \textit{where-learning}. Each action demonstrated to a simulated learner can  have multiple explanations each with different bindings. Consequently \textit{where-learning}, like \textit{how-learning}, can benefit from being provided \textit{foci-of-attention} to make the binding that correctly explains a demonstration unambiguous.

\subsection{When-Learning Mechanism}
\subsubsection{Computational-Level}
The \textit{when-learning} mechanism learns the \textit{when-part} of a skill, a binary function (often in the form of a set of conditions) that determines whether or not a particular candidate skill application should fire given the current state of the tutoring system. While \textit{where-} and \textit{how-learning} can typically learn only from positive examples, \textit{when-learning} learning requires both positive and negative examples since it must learn a \textit{when-part} for each skill that accepts correct candidate skill applications and rejects incorrect ones.  Additionally, since a \textit{when-part} has to be able to accept or reject a wide variety of candidate skill applications, \textit{when-learning} must be able to produce \textit{when-parts} that generalize across similar situations among many training examples. Broadly speaking generalization in \textit{when-learning} can be achieved in two ways: 1) include a preprocessing step in \textit{when-learning} to alter the representation of the state prior to passing it to the core \textit{when-learning} mechanism so that the state includes contextual information about each candidate skill application, and 2) design the \textit{when-learning} mechanism so that it can invent or restructure features present in the state to help it generalize across examples.

%\textit{When-learning} also typically takes more learning opportunities to converge than \textit{how-} and \textit{where-learning} since several positive and negative examples must be collected to determine which of the many features present across different interface states are necessary and sufficient conditions for applying each skill.

%In general, a \textit{when-learning} mechanism should produce any sort of binary classifier that can take in the tutoring system state, and a candidate skill application and output whether or not the candidate skill application should be applied.
%Consequently, \textit{when-part} conditions typically take on more diverse forms than the \textit{where-part}. \textit{When-part} conditions can evaluate both conjunctive and disjunctive predicates, establish threshold requirements on continuous values, or involve even more complex functions on the problem state.

%Unlike the \textit{where-part}, which generates bindings for a particular set of interface elements, \textit{when-learning} learns conditions on the entire  state of the interface

\subsubsection{Algorithmic-Level: State Pre-Processing}
\begin{figure}
    \centering
    \includegraphics[width=.35\linewidth]{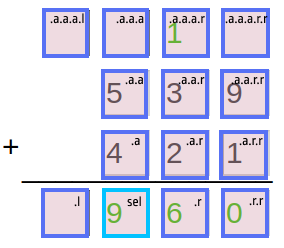}
    \caption{The state of the tutoring system renamed relative to the selection interface element of a candidate skill application (i.e., sel). '.a', '.b','.l','.r' are short-hands for '.above', '.below', '.left', '.right'.}
    \label{fig:relative}
\end{figure}

AL implements a few methods for state pre-processing that can be integrated with a \textit{when-learning} mechanism to help generalize across examples. One method, 'AppendBinding' simply adds the selection and arguments of the candidate skill application's \textit{binding} to the state. However, the method used by default, called 'Relative', takes this a step further and additionally relabels the entire state of the tutoring system relative to the binding's selection. This 'Relative' prepossessing method allows AL's various \textit{when-learning} mechanisms to utilize features of the interface surrounding a candidate action instead of just considering the absolute features of the interface. Figure \ref{fig:relative} shows this process applied to our running multi-column addition example. For instance, the \textit{when-parts} of the skills associated with placing the ones digit of partial sums would probably check that the interface "sel.right", (i.e. the one to the right of the selection), is filled in and locked, since each column must be handled in a right to left fashion. 

%Another form of state augmentation that the Apprentice Learner framework utilizes for \textit{when-learning}, is relabeling features of the tutoring system state relative to the selection interface element of a candidate skill application. Figure \ref{fig:relative} shows an example of this relabelling. By considering interface elements relative to the selection, the agent can reason about whether a skill should be applied in the context that it would take place. This way the agent can generalize across different candidate applications of a skill to find a set of conditions that work in all cases. All literals that occur in the the resulting \textit{when-part} conditions would then look something like '?sel.above.above.above.contentEditable == False', where interface elements that show up in conditions are located by de-referencing adjacency features (like '.above') starting with the '?sel' interface element. 

\subsubsection{Algorithmic-Level: Classifiers}
Several \textit{when-learning} mechanisms have been used across Simstudent and AL. One of the most common \textit{when-learning} mechanisms used in AL is the decision tree algorithm. The decision tree algorithm \cite{quinlan_C45} finds sets of conditions that maximally filter data into bins of items that share the same class label. In the case of \textit{when-learning} a decision tree learns conditions that separate positive training examples from negative ones. By recursively choosing features that optimally split the data into more purely positive and negative groups a tree of conditions is formed that filters the training examples into a number of purely positive or negative leaf nodes (if possible). The path from the root of the decision tree to any purely positive leaf node constitutes a conjunction of logical statements that accept only positively labelled candidate skill applications among those experienced so far. The disjunction of these conjunctions constitute the \textit{when-part} conditions.

FOIL (First-Order Inductive Learner) \cite{quinlan_FOIL} is an inductive logic programming based classifier that is used as a \textit{when-learning} mechanism in SimStudent. Instead of using a divide and conquer strategy FOIL uses the principle of sequential covering. Conjunctions of literal terms are built up until the conjunctions select only positively labelled instances, and then the selected instances are removed from the training set and the process is repeated until all positive instances are covered. Both the decision tree algorithm and FOIL are symbolic methods and can output a set of conditions that can be  expressed in disjunctive normal form. However unlike the decision tree, FOIL is typically given a target predicate with a number of variable arguments, and considers possible ways of using these argument variables with known predicates given to it as background knowledge. 

TRESTLE \cite{maclellan2016trestle} is another \textit{when-learning} mechanism available in AL is, which differs from the decision tree and FOIL in that it is an incremental algorithm---it learns from one example at a time instead of needing to be repeatedly refit. This makes TRESTLE a somewhat more cognitively plausible \textit{when-learning} mechanism than decision trees or FOIL since humans certainly don't incorporate new information by comparing it against a perfect memory of every other example they have seen thus far. TRESTLE is essentially an extension of the COBWEB \cite{fisher1987knowledge} category learning algorithm that includes structure mapping---before incorporating a new state, or when predicting the legality of a skill application, it tries to find a relabelling of the state that best fits the concepts that it has already learned. 

%A common thread among \textit{when-learning} mechanisms is the need to generalize across similar candidate applications of a skill. As we mentioned in the previous subsection relabelling the state relative to the selection of a candidate skill application is one way to achieve generalization across positively and negatively labelled examples of a skill being used. However, 
FOIL and TRESTLE  both have built in strategies for turning an otherwise grounded (i.e. attached to a particular instance) example into one that can be re-expressed with variables. Like pre-processing, these capabilities can help \textit{when-learning} generalize across examples. Future research may seek a more detailed understanding of how and when these built in generalization capabilities are beneficial beyond what can be afforded by the extra features provided by pre-processing.

%\subsubsection{Algorithmic-Level: Cross Skill Inference}
Another feature which can be mixed into \textit{when-learning} is the inclusion of \textit{implicit negatives}; when one skill receives a positive example all other skills receive implicit negative examples \cite{matsuda2005building}. Implicit negatives implement the intuition that if a particular action is correct at a step then it is probably the only correct action at that step. Problems that have multiple solutions complicate this form of inference, since it is not strictly true that there is only one correct action at each step. But, if the goal is to model the learning of mastery behavior, and not knowledge of all solution paths (like in authoring use cases \cite{weitekamp2020CHI}) implicit negatives can help \textit{when-learning} learn to reject alternative actions and converge to appropriate mastery behavior with fewer instances of explicit negative feedback.

%Unlike normal negative examples, implicit negatives can be overridden by positive examples that contradict them. For example, ... In attempts to re-implement this behavior in AL we have found that implementing an effective means of overriding  false implicit negatives is challenging, since an implicit negative will be overridden only if the state that it occurred in occurs again but in the presence of a positive example that didn't occur the first time. In the course of training with random problems this essentially never happens. This is problematic in cases where a set of implicit negatives support one action "A" of two possible actions ["A", "B"], but in a later problem only "B" is legal. In this case the implicit negative may trip up \textit{when-learning} by contradicting otherwise positive evidence in support of a \textit{necessary} legality condition and cause \textit{when-learning} to overspecialize.  

\subsubsection{Instructional Affordances}
Like \textit{how-} and \textit{where-learning}, \textit{when-learning} benefits from extra instructional information like skill labels and \textit{foci-of-attention} that aide the simulated learner in finding the correct explanation for an action demonstrated to it. Having the correct explanation for an action reduces the chance that it is attributed to the wrong skill, leading spurious examples to be injected into the \textit{where-} and \textit{when-learning} mechanisms' training sets \cite{weitekamp2021toward}. Certain agent configurations are more sensitive to the misattribution of feedback, for example if \textit{implicit negatives} are enabled then a misattributed positive example can have wide ranging effects.

Additionally, although it has not been implemented in a simulated learner, the programming by demonstration system GAMUT \cite{mcdaniel1997gamut}, implements a feature similar to \textit{foci-of-attention} where the system can be given hints about important interface features to condition on. In principle, the same interaction could be used as hints to a simulated learner's \textit{when-learning} mechanism.

\subsection{A Conceptual Comparison With Reinforcement Learning} \label{RL_analogy}

%In reading the following sections, which go into each of DIPL's learning mechanisms in detail, 

%Readers familiar with reinforcement learning should keep in mind that in addition to using several learning mechanisms agents that embody the DIPL must 

%the DIPL framing of procedure learning, tasks the agent with learning both how to produce actions and how to choose between them, whereas reinforcement learning, in domains with discrete actions, only learns the latter. This is one reason why the \textit{how-} and \textit{where-learning}  mechanisms are necessary.

Readers familiar with reinforcement learning (RL) may benefit from the following conceptual comparison between DIPL and RL. RL agents are typically given a set of candidate actions predetermined by their environment and learn a policy $\pi(a|s)$ that determines the probability that an action $a$ will be taken given a particular state $s$ \cite{sutton2018reinforcement}. DIPL simulated learners by contrast use prior knowledge of basic domain general functions to compose new domain specific functions that produce actions. RL policies are somewhat analogous to \textit{when-learning} in DIPL, since both RL policies and \textit{when-parts} help solve the problem of deciding which of several candidate actions ought to be taken to exhibit a target behavior. But, DIPL additionally uses it's \textit{where-} and \textit{how-learning} mechanisms to learn how to produce actions. Expressed as a function, an RL agent takes in a state $s$ and \textit{chooses} an action with it's policy $F_{\pi}(s) \rightarrow a$. By contrast, in DIPL \textit{where-learning} produces a generative function, the \textit{where-part}, to produce \textit{candidate skill applications}, and \textit{when-learning} produces a binary function, the \textit{when-part}, to determine if those candidate skill applications should be taken. Additionally, instead of having a fixed space of actions, \textit{where-learning} and \textit{how-learning} (which learns the \textit{how-part}) learn the space of possible actions. Thus DIPL's effective policy is decomposed across several functions that are learned by different learning mechanisms and has the form $When(s,Where(s)) \rightarrow How(Where(s))=a$.

\section{A Comparison With Alternative Methods} \label{results}

%In section \ref{?} we outlined several benefits to building simulations around the DIPL theory especially with regards to supporting an efficient learning rate (\ref{C:lrnrate}) and a diversity of instructional interactions (\ref{C:inst}).
In this section we empirically support the claims that the DIPL computational-level theory is well suited to achieving human-level learning efficiency (\ref{C:lrnrate}) and task performance (\ref{C:perf}) in step-based academic problem solving. In the following experiments we wish to test empirically several claims about the DIPL computational-level theory relative to a few alternatives including deep-learing based reinforcement learning methods, symbolic methods used outside of the context of the DIPL theory, and alternate specifications of the DIPL theory that use fewer mechanisms.

By testing the learning rates of these various competing methods we hope to support the following claims: A) Connectionist back-propagation based learning methods require considerably more training than symbolic methods and thus are not well suited to achieving human-level learning rates (\ref{C:lrnrate}). B) Decomposed inductive procedure learning (i.e. DIPL) is more learning efficient than using a single learning mechanism. C) all of DIPL's three core learning mechanisms must be present to achieve a human-level learning rate. In particular it is beneficial to have a \textit{how-learning} mechanism to learn skill RHSs, and LHS learning benefits from being decomposed into \textit{where-} and \textit{when-learning}.

%There is a difference in learning rates between decomposed learning mechanisms whereby multiple learning mechanisms interact to acquire different pieces of a single goal, and the use of separate but independent learning mechanisms. D) The learning rate of AL agents that embody DIPL and use \textit{where-} and \textit{when-learning} to produce skill LHSs have a faster learning rate than AL agents that do not embody DIPL and use just a single learning mechanism to produce LHSs.

\subsection{Methods}
To test these claims we compare the performance per learning opportunity between five machine learning agents on the two ITSs outlined in section \ref{procedure}, Multi-Column Addition and Fraction Arithmetic. We have chosen these two math domains because they are possible to learn without considerable prior knowledge, but relatively challenging domains to learn from scratch since they each embody almost all of the complexity factors we outlined in section \ref{complexity}, and unlike many of the linguistic and scientific domains that simulated learners have learned in prior work \cite{maclellan2020domain} \cite{matsuda2015teaching}, these two math domains can be adapted so that they can be tractably learned by deep reinforcement learning, since thousands of problems can be generated for these two domains with a small finite set of symbols.

Models 1-3 frame task learning, as an Reinforcement Learning agent would, as a \textbf{state-action response} problem: the problem is framed as selecting between all possible SAIs given the one-hot encoded problem state. This framing of task learning is implemented using the 'TutorGym' setup outlined in \cite{maclellan2021RL}. In this setup the starting fields can take values in the range 0-9 for multi-column addition and 1-15 for fraction arithmetic. The set of SAIs is the set of all selections and integers within a feasible output range (0-9, and 1-450, respectively). Since Models 1-3 have no means of asking for demonstrations, they are automatically provided the correct next action after each single incorrect attempt and are moved on to the next step of the problem.

Models 4-5 frame task learning, as AL and SimStuent do, which we'll call a \textbf{state-skill response} problem going forward. By contrast to a state-action response problem, in a state-skill response no preset actions are given, the agents must induce skills to explain demonstrations, and may manipulate the state representation as necessary by its different learning mechanisms. These models are given relevant domain-general prior knowledge including the ability to add, subtract, multiply, and divide, extract ones and tens digits, and recognize the equality of values in the interface. Like Models 1-3 these two models are provided correctness feedback in response to attempted actions, but are only provided demonstrations when requested. No additional instructional information like \textit{skill-labels} or \textit{foci-of-attention} are provided.

Ordered roughly according to our predictions of slowest to fastest in terms of per-opportunity learning rate the five models are:

\begin{enumerate}
    \item \textbf{Deep reinforcement learning}: (\textit{state-action response}) The default DQN setup provided by the stable-baselines library \cite{stable-baselines} is used. An initial exploration rate of 45\% is reduced to zero over the first 10\% of problems. In both domains three fully connected hidden layers are used with widths 200, and 800 for multi-column and fractions respectively. Training and model parameters were tuned via grid search. 
    \item \textbf{Single decision tree}: (\textit{state-action response}) Scikit-Learn's \cite{scikit-learn} decision tree classifier is used to predict the SAI given a one-hot encoded state.
    \item \textbf{Double decision tree}: (\textit{state-action response}). Like single decision tree except that one decision tree learns the selection (S) and another learns the action-type (A) and input (I) for the action's SAI.
    \item \textbf{Single-mech. LHS AL agent}: (\textit{state-skill response}) A variation of a typical AL agent that merges \textit{where-} and \textit{when-learning} into a single mechanism. This agent performs \textit{how-learning} as usual, but for the LHS, instead of using a \textit{where-part} to generate bindings for candidate skill applications and a \textit{when-part} to verify them, a single best binding is chosen per skill at each step by a decision tree. The decision tree treats the choosing of a binding as a multi-class prediction problem on the current interface state, where the set of classes are all of the bindings observed so far plus an option to produce no binding (i.e. not consider applying that skill). 
    \item \textbf{DIPL AL agent}: (\textit{state-skill response}) A typical AL agent, using the 'AntiUnify' implementation of \textit{where-learning} \cite{weitekamp2020CHI} and 'DecisionTree' implementation of \textit{when-learning}.
\end{enumerate}

100 agents were run for Models 4 and 5 for as many problems necessary for the agents to achieve an average error rate of less than 1\%, up to a maximum of 1000 problems. Due to constraints on time and computing resources (each decision tree model took nearly a week to run) just 6 agents were run for models 1-3 on 30,000 randomly generated problems. 

\subsection{Results}\label{results_results}

\begin{figure}
\begin{minipage}{.5\textwidth}
  \centering
  \includegraphics[width=\linewidth]{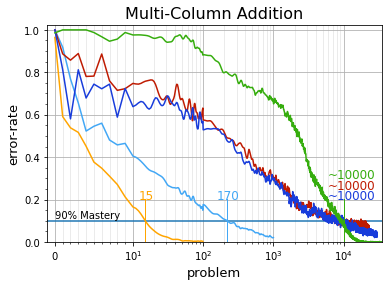}
  %\captionof{figure}{A figure}
  %\label{fig:test1}
\end{minipage}
\begin{minipage}{.5\textwidth}
  \centering
  \includegraphics[width=\linewidth]{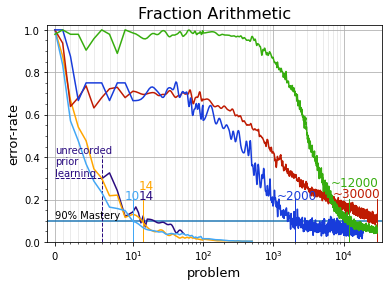}
  %\captionof{figure}{Another figure}
  %\label{fig:test2}
\end{minipage}
\begin{minipage}{\textwidth}
  \centering
  \includegraphics[width=\linewidth]{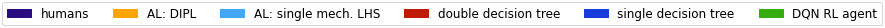}
\end{minipage}
\caption{Error-rate by problem (problem axis is log-scaled) plotted for our five agent models run on two math domains, multi-column addition and fraction arithmetic. Fraction arithmetic also includes human data offset by 6 problems (the point at which the initial error rate intersects the DIPL AL model). This is to account for learning from prior instruction before the data was collected. The 90\% mastery problem-intercept is graphed for each agent model, indicating the number of randomly selected problems that on average an agent must practice on to achieve a 10\% error rate. Piece-wise Gaussian smoothing is applied with increasing $\delta$ in each interval (0-$10^1$,$\delta=0$) ,($10^1$-$10^2$,$\delta=2$), ($>10^2$,$\delta=10$)}
\label{fig:results}
\end{figure}

%\begin{figure}
%\begin{center}
%  \includegraphics[width=0.8\linewidth]{figures/JAIED2021_results.png}
%  \caption{}
%  \label{fig:results}
%\end{center}
%\end{figure}

Figure \ref{fig:results} shows the average agent error rate per problem with the problem number axis graphed in log scale. We compare learning efficiency in terms of the number of problems needed to reach a common point of mastery, defined as an error rate below 10\%. In both the multi-column addition and fraction arithmetic domains, the deep reinforcement learning models (in green in Fig \ref{results}) and decision tree models (in red and dark blue) achieve mastery after thousands to tens of thousands of problems.  By contrast the DIPL style AL agents reach this point within 15 or fewer problems, and the single LHS AL model (light blue) was comparable to the DIPL AL agent in Fraction Arithmetic, but significantly slower in multi-column addition, reaching mastery at 170 problems. If we establish a reasonable cut-off of 50 problems as a plausible number of problems that a human might actually practice\footnote{this is twice as many problems as any student completed in the Fraction Arithmetic dataset} then we find that the DIPL AL agents are the only agents that can a achieve 90\% mastery within this range across both domains and continue to improve up to and past this point.

In the fraction arithmetic domain we have compared the learning rates of our various models to human data available on PSLC Datashop \cite{koedinger2010data} from a tutoring system study with fourth grade participants\footnote{https://pslcdatashop.web.cmu.edu/Project?id=243.}. The human learning curve is shown for Fraction Arithmetic in Figure \ref{results} as a dark purple line \footnote{we were unable to locate an appropriate comparison dataset for MultiColumn Addition}. Unlike human students our models begin practice without any prior instruction meaning they start with a 100\% error rate. However, many human students have already received some prior demonstrations and practice with feedback from teachers, parents, peers, and self study. As in prior work \cite{weitekamp2020investigating} we account for this discrepancy by displacing the human data to the point where the first opportunity human error rate matches that of the simulated learners. This displaced human learning curve crosses the line of mastery at the same point as the DIPL AL agents, and shows similar learning gains thereafter.

With some caveats, these results support our three claims. Contrary to our predication (A) the deep reinforcement learning model  required about the same number (not a larger number) of practice problems than the decision tree models to reach task mastery. It did, however, require orders of magnitude more practice than the DIPL and single LHS AL models as predicted. Along the way to mastery (a 10\% error rate) the deep learning consistently maintained a considerably higher error rate than any other model. 

These results additionally support the claim (B) that DIPL AL agents have a considerably faster learning rate than single mechanism methods. Recall that our single decision tree model differs from the DIPL AL agents in that it uses a single mechanism instead of DIPL's three, and that it must consequently frame learning as the deep reinforcement learning agent does, as a \textit{state-action response} problem instead of a \textit{state-skill response} problem, since without a \textit{how-learning} mechanism to produce RHSs it cannot learn distinct skills.

These data provide additional insights into what it is about DIPL that allows it to learn academically relevant tasks efficiently. Firstly, although these data certainly indicate that DIPL's particular manner of decomposing learning is beneficial, they also show that it is not necessarily the case in general that having multiple mechanisms is beneficial. The double decision tree model uses two mechanisms to learn the same tasks as the single decision tree model, yet their is no discernible difference between the two in the multi-column addition case and, in the fraction arithmetic case the double decision tree requires more than an order of magnitude more training to reach mastery (30000 versus 2000 problems). A key difference to keep in mind between DIPL and the double decision tree is that DIPL's learning mechanisms operate in a cooperative manner, while the double decision tree model is simply using two mechanisms to produce two parts of an SAI (the selection and input) independently of one another. This point is elucidated in Section \ref{theory}.

Finally, two observations from these data support claim (C), that all three of DIPL's core learning mechanisms help it achieve a human-level learning rate. First, the single LHS AL model (light blue in Fig. 7) which includes a \textit{how-learning} mechanism is orders of magnitude more efficient than models 1-3 which do not have \textit{how-learning} mechanisms. Second, in the Multi-Column Addition domain the AL DIPL model (gold in Fig. 7) which has all three core mechanisms is more than an order of magnitude more efficient (15 versus 170 problems) than the single LHS AL model in which \textit{where-} and \textit{when-learning} are combined into a single learning mechanism. Thus these data support both the framing of task learning as a matter of acquiring skills that are each decomposed into RHSs (how-parts) and LHS, and the further decomposition of LHSs into separately learned \textit{where-} and \textit{when-parts}.

\subsection{Discussion}

Overall these results support the claim that the DIPL computational-level theory is well suited to simulating the efficiency of human learning. In section \ref{theory} we support these results further with theoretical justification. However, a few matters are worth discussing further in reference to this data. 

Firstly, in the fraction arithmetic domain the single mechanism LHS model performed comparably to the AL DIPL model, but was significantly slower than the AL DIPL model in the multi-column addition domain. The advantage of maintaining the \textit{where}/\textit{when} decomposition in the multi-column addition domain is that having a separate \textit{where-learning} mechanism helps \textit{when-learning} generalize across repeated uses of skills in different locations in the problem display. By contrast, the lack of a \textit{where-learning} mechanism in the single LHS AL model prevents effective generalization. In the fraction arithmetic domain however, there is a consistent pattern to where skills are used, and most skills are only used once per problem in one particular interface element. Thus, there is not a considerable difference in learning rate between these two models in the fraction arithmetic domain.

Secondly, it should be noted that while the deep reinforcement learning model's initially high error rate in both domains is partially explained by it's need to do some initial random exploration (it will fail to converge without this), the slow incremental nature of gradient descent plays a decidedly bigger role. Unlike the other four methods which all utilize some form of symbolic machine learning, the deep-learning model does not maintain a solution consistent with the training data collected so far. Rather at every training opportunity the current weights of the deep reinforcement learning model are incrementally different than the weights at the opportunity before it, and represent just one weak hypothesis in an immense continuous topological space of possibilities. 

Consequently, we doubt that deep learning will be a viable tool for building agents that learn academically relevant domains from scratch as readily and with as few examples as our DIPL AL agents. Connectionist representations of knowledge learned via gradient descent are certainly well suited to learning tasks that have noisy high-dimensional stimuli, and complex continuous response patterns. But, natural language tasks notwithstanding, academically relevant domains are almost always noiseless and symbolically expressible as logical rules. Thus, symbolic machine learning methods have a distinct advantage. By embodying a cognitive bias for knowledge representations consistent with the domain in question they are able to make large leaps of induction from single examples. As our comparison with human data shows, these large inductive leaps produce reductions in task error consistent with human learning efficiency.

\section{Benefits of Decomposing Learning into Multiple Mechanisms} \label{theory}

In the following sections we outline a theory to explain why DIPL's particular manner of decomposing learning benefits per-opportunity learning efficiency (\ref{C:lrnrate}). We establish why the learning tasks tackled by each learning mechanism are substantially different from one another. We consider how each learning mechanism establishes, narrows, and selects from a space of solutions, how these spaces of solutions interact with one another, and what inductions and inductive biases can be selectively made by treating them separately. 

\subsection{The Nature of RHS and LHS Solution Spaces Differ}
In section \ref{results_results} the largest gains in learning efficiency were established by introducing a decomposition between RHS learning (\textit{how-learning}) and LHS learning (\textit{where}/\textit{when-learning)}, as shown by the difference between the state-action response conditions and the state-skill response conditions. To understand the benefit of this decomposition it is helpful to think about the space of all skills as the Cartesian product of the set of all RHSs  $\mathcal{R}$ and the set of all LHSs $\mathcal{L}$. From this perspective learning skills is, in a broad conceptual sense, a matter of leveraging evidence from instructional interactions to search within the spaces $\mathcal{R}$ and $\mathcal{L}$ for combinations of RHSs and LHSs consistent with what has been observed so far. 

%as shown by the contrast between AL sing mech LHS and the decision tree. Using the notation developed above, we can characterize this distinction as When(State)->Action (RL without decomposition) versus When(State)->How(State)=Action.  

It is, however, important to appreciate that the spaces $\mathcal{R}$ and $\mathcal{L}$ are very different in nature, and require different kinds of search to be navigated efficiently. Each element in the space of RHSs $\mathcal{R}$ is a composition of domain-general functions that produce actions, whereas each element in the space of LHSs $\mathcal{L}$ is a set of logical conditions that accept or reject possible skill \textit{bindings}\footnote{Recall we are limiting our consideration here to symbolic methods.}. Within both of these spaces there is some desirable subset of elements consistent with the evidence collected so far. 

For $\mathcal{L}$ this subset has some structure and boundaries. The elements of $\mathcal{L}$ can be altered and be made more general to match more potential state-binding pairs or more specific to match fewer. Logical condition learning algorithms like decision trees and FOIL grow an effective set of conditions by heuristically guided search through the space of possible conditions to find a set of conditions consistent with training data. Each step in this process brings a single partially consistent solution closer to labelling all positive and negative examples correctly. In limited cases version-spaces \cite{mitchell1982generalization} can even characterize the entirety of the space of consistent conditions by finding two boundary sets, the sets of elements that can be made no more general or admit some negative examples, and no more specific or reject some positive ones. 

%Some elements can be made no more general or admit some negative examples, or no more specific or reject some positive ones. In constrained cases version-spaces can even find the boundary elements that enclose all consistent solutions \cite{mitchell1982generalization}.

For $\mathcal{R}$ on the other hand there is not any analogous means of characterizing or bounding a consistent region of solutions. And more importantly, there is no certain means of determining how an inconsistent element of $\mathcal{R}$ can be augmented to necessarily bring it closer to being consistent with a set of examples. A step through the space $\mathcal{L}$ involves incorporating new literals into logical formulae, and either makes measurable progress toward consistency or does not (i.e. more examples are correctly labelled as positive/negative). By contrast a step through $\mathcal{R}$ involves composing functions and produces no certain measure of progress. Most steps in a typical construction of $\mathcal{R}$ will produce compositions inconsistent with any example actions, even those that are on path to a consistent solution. Additionally partially consistent solutions are not necessarily closer to being consistent with all of the example actions. Consequently in the worst case finding a consistent element of $\mathcal{R}$ can mean searching through all possible function compositions up to a fixed depth. For this reason skill learning is most efficiently achieved by using different sets of mechanisms for RHS and LHS learning---each set suited to the unique characteristics of the spaces of solutions they must search through.

\subsection{Two Mechanism LHS Learning Allows for Multiple Inductive Biases}
%In section \ref{results_results} further gains in efficiency were achieved by decomposing LHS learning into \textit{where-} and \textit{when-learning}.
One benefit of decomposing LHS learning into \textit{where-} and \textit{when-learning} is that using two mechanisms enables a choice of two different inductive biases. One important choice of bias is the tendency for a learning algorithm to produce a classifier or set of conditions that accept either many examples beyond the positive examples in the training set or few. We'll call the extremes of this spectrum of biases \textit{general-solution bias} and \textit{specific-solution bias} respectively. Particular strategies for constructing conditions can produce these biases. A condition learning algorithm's process of constructing conditions can either progress from \textit{specific-to-general}, as in least-general generalization methods \cite{lgg} where very specialized logical statements are generalized to accept new examples, or \textit{general-to-specific}, where literals are conjoined with or otherwise specialized within a logical statement to exclude some negative examples\footnote{\textit{Specific-to-general} and \textit{general-to-specific} learning are often discussed in the context of version space construction algorithms \cite{mitchell1982generalization}, we are not invoking the specifics of these algorithms here, only patterns of condition construction.}. Decision trees applied in the context of \textit{when-learning} fall into this latter category since the effective conditions of the tree only specialize with each refitting\footnote{More specifically, when refit to incorporate some new examples the new tree will either be a specialization of the previous tree, or structurally different from the previous tree, typically with a similar number of nodes. The new tree will however not be a generalization (i.e. pruned version) of the previous tree since new data cannot make previously found splits unnecessary.}. In the context of these strategies \textit{general-} and \textit{specific-solution biases} arise from condition construction progressing via generalization or specialization and stopping on a condition within the region of consistent conditions. The tendency of an algorithm toward one of these biases can however be considered empirically even when these strategies are not used. In the context of simulated learners, a \textit{general-solution bias} in \textit{where-} and \textit{when-learning} will cause the agent to take more actions, whereas a \textit{specific-solution bias} will cause it to ask for more hints.

In prior work \textit{where-learning} has utilized a \textit{specific-to-general} learning strategy and in turn embodied a \textit{specific-solution bias}. This choice immensely cuts down on the quantity of candidate skill applications that \textit{when-learning} needs to subsequently build conditions to exclude. Deviating far from this choice of bias would compromise most of the efficiency benefits of decomposing LHS learning. If instead \textit{where-learning} utilized a \textit{general-to-specific} construction strategy and consequently a \textit{general-solution bias}, then a newly initialized skill it would produce a \textit{where-part} with no constraints and propose candidate skill applications for all permutations of interface elements in the interface. For $n$ interface elements and a skill with $n_{f_c}$ arguments, the \textit{where-part} could propose any of  $\frac{n!}{(n-(n_{f_c}+1))!}$ permutations of bindings. The agent would in turn try a large number of skill applications with selections and arguments that have no correspondence to any observed demonstrations, and would receive a great deal of negative feedback in response. Thus, this bias would put an unnecessary burden on \textit{when-learning} to exclude a great number of errant candidate skill applications, and perhaps take many hundreds of additional examples to build conditions that exclude all of the extraneous permutations produced by the over general \textit{where-parts}.

In past SimStudent and AL implementations, \textit{when-learning} has usually been implemented with learning mechanisms that have a more \textit{general-solution bias} in contrast to the specific-solution bias of \textit{where-learning}. There are several reasons to favor the \textit{general-solution} side. Some prior work has shown that when working on multi-step tasks in an ITS, humans tend to primarily ask for hints in the first few opportunities of practice and consistently attempt each step on subsequent opportunities; learning from the resulting correctness feedback \cite{weitekamp2020investigating}. \textit{When-learning} with a highly \textit{specific-solution bias} on the other hand curtails actions and can potentially produce more hint requests well into training since more specific \textit{when-parts} tend to reject skill applications in states that are slightly different than what the agent has encountered before. Additionally when simulated learners ask for more hints and generate fewer attempted actions their overall learning can suffer because they generate fewer opportunities to receive negative feedback. For example, Matsuda found that SimStudent learned considerably more from making mistakes and receiving negative feedback than from only observing positive examples in the form of demonstrations \cite{matsuda2015teaching}.  In essence, \textit{when-learning} with a general-solution bias produces more active learning \cite{settles2009active} opportunities whereby the system essentially elicits negative examples to narrow prior overly-general \textit{when-parts}.  

Another reason to favor a \textit{general-solution bias} from \textit{when-learning} is that more general \textit{when-parts} require less cognitive processing to execute. The most general possible \textit{when-part} would impose no constraints on acceptable candidate skill applications and thus require no processing at all. With negative examples, \textit{when-learning} adds conditions to the \textit{when-part} making it more specific and more cognitively taxing to process. These additions of extra processing/code are only introduced as needed. A specific-solution bias in \textit{when-learning} by contrast would yield more processing/code in early learning and reduce it in later learning. The specificity of \textit{where-parts} by contrast has little bearing on how cognitively demanding they are to process since they typically only impose conditions on the selection and arguments to generate new bindings.

\subsection{Learning Mechanism Decomposition Has a Clear Generalization and Evidence Attribution Structure}

Returning to our comparison between DIPL and RL based agents (section \ref{RL_analogy}), recall that while DIPL agents induce how-parts that produce different types/sets of actions, RL bases agents typically learn one global policy function that maps states to a set of pre-established actions. RL agents incorporate new evidence in the form of reward by reinforcing the activation of positive reward producing behavior, and suppressing the activation of negative reward producing behavior. One of the main challenges in reinforcement learning is to learn generalizations between similar (parts of) states and similar actions.

DIPL based agents solve these two generalization problems by using \textit{how-learning} to generalize across actions and  \textit{where-learning} to generalize across similar parts of states. When an agent observes new evidence, such as a demonstrated action or positive/negative feedback on an attempt, the evidence can be attributed to a particular subset of the agent's skills that are consistent with the corresponding action. For example, when an agent observes a demonstrated action, it will only attribute that action to skills with a \textit{how-parts} that could have produced the action. Additionally if foci-of-attention are provided with a demonstration, only candidate skill applications that utilize the same arguments (i.e., that have \textit{where-part} output that matches the given foci) receive positive feedback from that demonstration.

%One features of simulated learners implementing DIPL is that demonstrations, and
 The decomposed structure of skills into  \textit{where}, \textit{when}, and \textit{how-parts} facilitates further efficiency in distributing the reward signal within skills. Positive and negative feedback are each used in different ways, and are attributed to different learning mechanisms in a manner that reduces the ambiguity of how those different kinds of evidence should be used for learning. At the core of this idea is the fact that \textit{how-} and \textit{where-learning} only learn from demonstrations, and in some circumstances positive feedback, but \textit{when-learning} is the only learning mechanism that utilizes all forms of feedback. Negative feedback is only used by \textit{when-learning}, and is essential to establishing which of many possible sets of conditions are the \textit{necessary} and \textit{sufficient} conditions for a legal application of a skill. \footnote{One caveat is that in some instances negative feedback can force the agent to ask for a demonstration which may trigger new how-learning}
%\textit{How-learning} does not utilize negative feedback since it is exclusively driven by demonstration, and \textit{where-learning} only needs positive examples to establish a generalization between them.

This attribution structure stands in contrast to the "end-to-end" learning in connectionist deep-learning models, often used in RL, whereby rewards guide collective loss function optimization. This end-to-end use of rewards has a desirable simplicity. However, gradient descent based learning typically requires many more examples (contravening \ref{C:lrnrate}), and the resulting functionality is typically difficult to explain, since knowledge is distributed across thousands of weights. By contrast DIPL's decomposed mechanisms allow simulated learners to be built with the human-like capability to acquire somewhat accurate and explainable partial skills (mostly by explaining demonstrations), and over the course of a handful of problems, refine those not quite perfect skills, through error feedback, to a point of mastery. \textit{How-} and \textit{where-learning} are the driving forces in the first few problems when skills are still being induced and \textit{when-learning} is the primary engine of skill refinement.

\subsection{Decomposed Learning Mechanisms Are Efficient Because They Share Information}
An important distinction can be drawn between what we are referring to as \textit{decomposed} learning mechanisms and \textit{separate} learning mechanisms. What we mean when we say DIPL learning mechanisms are "decomposed" is that the learning mechanisms are split up into parts that collectively solve a larger learning problem: the acquisition of functionally accurate skills. This means that the inductions of one learning mechanism can help guide the inductions of other learning mechanisms, or if we are to consider learning as search, one learning mechanism can constrain the space of solutions for another learning mechanism. By contrast mechanisms that are completely separate learn independently from one another and thus cannot inform each other's learning. Since DIPL concretely establishes the role of each learning mechanism and their relationships to each other, different forms of instructional interactions can be directed to particular learning mechanisms unambiguously. It is easiest to follow this line of reasoning by continuing with mutli-column addition as a running example:

\begin{center}
%\centering{
    \begin{tabular}{cccc}
      &  & 2 & 7 \\
    + &  & 3 & 5 \\
    \hline
      &  &  \\
     \end{tabular}
     \quad\quad
     \begin{tabular}{cccc}
      &  & 1 & 7 \\
    + &  & 9 & 6 \\
    \hline
      &  &  \\
     \end{tabular}
\end{center}    
%}

\subsubsection{Example}
Consider a fresh simulated learner presented the following two multi-column addition problems. This simulated learner has prior knowledge of 7 domain-general functions in its function set $F$: Add(x,y), Add3(x,y,z) Subtract(x,y), Divide(x,y), Multiply(x,y), GetTensPlace(x), GetOnesPlace(y). The state has $n=4$ numerical values, and to find the ground truth \textit{how} function compositions the agent must search out to at least a depth of $D=2$. In general before an agent has seen any demonstrations the space of possible \textit{how-parts} that it might induce is of the order $O(|F|^{2^D-1}n^{2^D})$, so in this example around $7^{3}4^{4} = 87808$. The agent is shown the first step of the left problem (placing a 2 below the right-most column), there are 130 possible compositions of these 7 functions to depth 2 that produce the value '2' given the four numbers in the interface. These include simply copying the 2, and subtracting 5 from 7. If foci-of-attention were given with this demonstration then there are only 20 compositions that explain the example instead of 130. If the example was accompanied by a natural language description that included phrases such as "add", and "one's place" then an agent could conceivably impose further restrictions, although this functionality has yet to be implemented. Additionally if skill labels are given as well on a demonstration of this step in the the next problem (place a 1 below the right-most column) then the intersection of function compositions that explain both examples reduces to 1. 

\subsubsection{The \textit{How-Where} Connection} The relatively small number of \textit{how-compositions} (130 or 20 if foci are used) out of the total possibilities (\char`\~87000) constrains the problem of finding an initial \textit{where-part} generalization considerably. Of these the 130 possibilities only 27 ordered subsets of arguments (or just 2 if foci are used) are plausible out of 64 possibilities. \textit{How-learning} then in effect seeds \textit{where-learning}. For each of these 27 (or 2) possibilities one most-specific set of \textit{where-part} conditions is implied. Each of these single most-specific hypotheses establish a space of possible generalizations of a size $(2^{2*7})= 16384$ (assuming 2 arguments and 7 features per interface element). However, the size of this space is of little consequence since the correct generalization would be achieved with just one other positive example of the skill being applied in another column. 

%Since the evidence of the ground truth legality conditions are somewhat more ambiguous, 

It should be clear, so far, from this example that despite contending with an initially very large hypothesis spaces, the forms of induction used in \textit{how-} and \textit{where-learning} are aggressive, excluding a large number of possibilities quickly, and can be expected to converge to their final forms in the course of just a few examples. \textit{When-learning}, on the other hand, proceeds somewhat more slowly than the other two mechanisms, since it must see many examples to determine which features only occur in positive examples and which ones co-occur only by coincidence from examples seen so far. This pattern is consistent with studies of humans which have established that it is often harder to learn "when" to do something than it is to learn how to do it \cite{chi1981categorization} \cite{zhu1996cue}. 

\subsubsection{The \textit{Where-When} Connection} As indicated by the difference in section \ref{results} between the AL DIPL and AL single LHS models in multi-column addition. \textit{Where-learning} can speed up \textit{when-learning} in some instances and help it to generalize across skill applications. Conceptually a \textit{where-learning} serves a similar role as convolution filters and attention mechanisms in neural networks. \textit{Where-parts} pick out common patterns of features so that features exhibiting a common pattern can be grouped and exposed to similar downstream processing steps (in this case \textit{when-parts}). Specifically, in DIPL \textit{where-learning} reframes the search for \textit{necessary} and \textit{sufficient} legality conditions around candidate bindings. This reframing is achieved either by relabelling the state (as with AL) or treating the binding values as variables (as with SimStudent). %The factor by which \textit{where-learning} speeds up \textit{when-learning} is approximately a factor of the average number of instances of a skill per problem instance. 
In addition since \textit{where-learning}'s generalization capabilities allow an agent to apply learned skills in unseen situations, the interaction between \textit{where-} and \textit{when-learning} may enable agents to transfer knowledge between interfaces and tasks. This has however, not been demonstrated empirically. %\textit{Where-part} generalization can also produce the human-like error of misapplying skills between situations that have surface level similarity (\ref{C:errors})[]. 

\section{Conclusion} \label{conclusion}

In this work we have argued for a theoretical investigation of human learning by constructing simulated learners that are capable of learning educationally relevant tasks such as mathematical or scientific procedures at a rate similar to human learners. We have characterized past work on simulated learners over the last 16 years through the lens of Marr's three levels of evaluating cognitive capacities. At the computational-level we presented the Decomposed Inductive Procedure Learning (DIPL) theory which describes the shared computational-level structure of these works and we discussed different algorithmic-level implementations of DIPL's core learning mechanisms made by SimStudent and implementations of the Apprentice Learner (AL) framework. Toward evaluating our DIPL theory we have identified six capacities that a simulated learner should be able to achieve. In summary, a simulated learner should be able to perform as accurately as a human learner (\ref{C:perf}) after experiencing the same number  (\ref{C:lrnrate}) and types (\ref{C:inst}) of learning opportunities as a human learner in an environment similar to the materials that would be presented to a human learner (\ref{C:mat}). Throughout this process the simulated learner should produce errors similar to those produced by human learners (\ref{C:errors}), and employ the same general methods of reasoning as a human learner (\ref{C:meth}). In this work we have explicitly addressed how agents implementing the DIPL theory address \ref{C:perf}-\ref{C:inst}, pointed to prior work that addresses \ref{C:errors}, and focused on induction (\ref{C:meth}.a), but have left further integration of deductive methods (\ref{C:meth}.b) to future work.

In this work we have placed a special emphasis on demonstrating the per-opportunity efficiency (\ref{C:perf}) of agents implementing the DIPL theory. Toward this end we have compared the per-opportunity performance of AL agents implementing DIPL on two math based intelligent tutoring systems, and compared it to the per-opportunity performance of several alternative machine learning methods (Section \ref{results}). This comparison revealed that DIPL-based AL agents had the fastest per-opportunity learning rate, and exhibited similar per-opportunity performance to humans in one domain where human data is available. Several additional results came from this comparison. Firstly representing procedural task learning in these domains as state-action response problems, the typical framing of reinforcement learning, is decidedly less efficient than methods that incorporate some \textit{how-learning} to form general skills that construct correct actions across multiple contexts. Secondly, within the methods that framed task learning as a state-action response problem, deep-reinforcement learning exhibited by far the highest error rate for most of training. As a result we have found it unlikely that deep-reinforcement learning, alone, will be a viable method for replicating human's per-opportunity learning efficiency on academically relevant tasks. Thirdly we found that combining DIPL's LHS learning mechanisms negatively impacted per-opportunity learning efficiency in certain types of domains, indicating that there is a benefit to including all of DIPL's three core learning mechanisms \textit{how-learning}, \textit{where-learning}, and \textit{when-learning}. Finally, we have presented further theoretical justification in section \ref{theory} for why DIPL's learning mechanisms exhibit high per-opportunity learning efficiency. 

Ultimately we find that the DIPL theory offers a unique set of capabilities and affordances that are amenable to simulating human learning on academically relevant procedural tasks. While many of today's popular machine learning approaches rely on large data sets of thousands to millions of examples, DIPL based agents are able to keep pace with human learning, requiring only a handful of learning opportunities.

\section{Acknowledgements}
The research reported here was supported in part by a training grant from the Institute of Education Sciences (R305B150008). Opinions expressed do not represent the views of the U.S. Department of Education.

\bibliographystyle{splncs04}
\bibliography{references}
%\bibitem{}
%\end{thebibliography}
\end{document}